\documentclass[submission]{eptcs}
\providecommand{\event}{AREA22}
\usepackage{underscore}
\usepackage{amssymb}

\title{Modelling the Turtle Python library in CSP}
\def\titlerunning{Modelling Turtle in CSP}

\author{Dara MacConville \and Marie Farrell \and Matt Luckcuck \and Rosemary Monahan
\institute{Department of Computer Science/Hamilton Institute, Maynooth University, Ireland \\
\email{dara.macconville.2018@mumail.ie}}
\thanks{This work has emanated from research conducted with the financial support of Science Foundation Ireland (SFI) under Grant Number SFI 18/CRT/6049}
}

\def\authorrunning{D. MacConville et al.}

\usepackage[xindy]{glossaries}

\newcommand{\csp}{\gls{csp} }

\newacronym{ltl}{LTL}{Linear-time Temporal Logic}

\newacronym{foltl}{FOLTL}{First-Order Linear-time Temporal Logic}

\newacronym{ctl}{CTL}{Computation Tree Logic}

\newacronym{ptl}{PTL}{Probabilistic Temporal Logic}

\newacronym{pctl}{PCTL}{Probabilistic Computation Tree Logic}

\newacronym{pctl*}{PCTL*}{Probabilistic Computation Tree Logic*}

\newacronym{tctl}{TCTL}{Timed Computation Tree Logic}

\newacronym{stl}{STL}{Signal Temporal Logic}

\newacronym{ltl-x}{LTL-X}{a version of Linear Time Logic without the `next' operator}

\newacronym{mtl}{MTL}{Metric Temporal Logic}

\newacronym{iactlk-x}{IACTLK-X}{a fragment of the temporal-epistemic logic CTLK, without the `next' operator}

\newacronym{mucalc}{$\mu$-calculus}{a strict superset of other temporal logics}

\newacronym{fotl}{FOTL}{First-Order Temporal Logic}

\newacronym{bdi}{BDI}{Belief-Desire-Intention}

\newacronym{prs}{PRS}{Procedural Reasoning System}

\newacronym{are}{ARE}{Autonomous Reasoning Engine}

\newacronym{dmars}{dMARS}{distributed Multi-Agent Reasoning System}

\newacronym{mas}{MAS}{Multi-Agent System}

\newacronym{asm}{ASM}{Abstract State Machines}

\newacronym{fsm}{FSM}{Finite-State Machine}

\newacronym{pfsm}{PFSM}{Probabilistic Finite-State Machine}

\newacronym{apfsm}{APFSM}{\textit{Autonomous} Probabilistic Finite-State Machine}

\newacronym{pha}{PHA}{Probabilistic Hybrid Automata}

\newacronym{fsa}{FSA}{Finite-State Automata}

\newacronym{ta}{TA}{Timed Automata}

\newacronym{gspn}{GSPN}{Generalised Stochastic Petri Net}

\newacronym{mopn}{MOPN}{Marked Ordinary Petri Net}

\newacronym{ba}{BA}{B\"{u}chi Automata}

\newacronym{dtmc}{DTMC}{Discrete-Time Markov Chain}

\newacronym{ctmc}{CTMC}{Continuous-Time Markov Chain}

\newacronym{pars}{PARS}{Process Algebra for Robot Schemas}

\newacronym{fsp}{FSP}{Finite State Processes}

\newacronym{piadl}{$\pi$ADL}{$\pi$ calculus combined with ADL}

\newacronym{csp}{CSP}{Communicating Sequential Processes}

\newacronym{cspm}{CSP$_M$}{machine-readable CSP}

\newacronym{hcsp}{HCSP}{Hybrid Communicating Sequential Processes}

\newacronym{shcsp}{SHCSP}{Stochastic Hybrid Communicating Sequential Processes}

\newacronym{ccs}{CCS}{Calculus of Communicating Systems}

\newacronym{wsccs}{WSCCS}{Weighted Synchronous CCS}

\newacronym{csp|b}{CSP$\|$B}{an integrated formalism combining CSP and B}

\newacronym{vhdl}{VHDL}{VHSIC Hardware Description Language}

\newacronym{adl}{ADL}{Architecture Description Language}

\newacronym{fdr}{FDR}{Failures-Divergences Refinement checker}

\newacronym{jpf}{JPF}{Java PathFinder}

\newacronym{ajpf}{AJPF}{Agent Java PathFinder}

\newacronym{mcmas}{MCMAS}{Model Checker for Multi-Agent Systems}

\newacronym{ltlmop}{LTLMoP}{Linear Temporal Logic MissiOn Planning}

\newacronym{cbmc}{CBMC}{C Bounded Model Checker}

\newacronym{psl}{PSL}{Property Specification Language}

\newacronym{fpga}{FPGA}{Field-Programmable Gate Array}

\newacronym{aadl}{AADL}{Architecture Analysis and Design Language}

\newacronym{lts}{LTS}{Labelled-Transition System}

\newacronym{uml}{UML}{Unified Modelling Language}

\newacronym{sysml}{SySML}{Systems Modelling Language}

\newacronym{robotml}{RobotML}{Robot Modelling Language}

\newacronym{dsl}{DSL}{Domain Specific Language}

\newacronym{sct}{SCT}{Supervisory Control Theory}

\newacronym{sat}{SAT}{Boolean satisfiability problem}

\newacronym{smt}{SMT}{Satisfiability Modulo Theory}

\newacronym{ide}{IDE}{Integrated Development Environment}

\newacronym{mde}{MDE}{Model-Driven Engineering}

\newacronym{ai}{AI}{Artificial Intelligence}

\newacronym{fdir}{FDIR}{Fault Detection, Isolation and Recovery}

\newacronym{ifm}{iFM}{Integrated Formal Methods}

\newacronym{ros}{ROS}{Robot Operating System}

\newacronym{amcl}{AMCL}{Adaptive Monte Carlo Localisation}

\newacronym{dl}{\text{\upshape\textsf{d{\kern-0.05em}L}}}{differential dynamic logic}

\newacronym{vm}{VM}{Virtual Machine}

\newacronym{bip}{BIP}{Behaviour Interaction Priority}

\newacronym{fol}{FOL}{First-Order Logic}

\newacronym{owl}{OWL}{Web Ontology Language}

\newacronym{ria}{RIA}{Restore Invariant Approach}

\newacronym{ocl}{OCL}{Object Constraint Language}

\newacronym{sil}{SIL}{Safety Integrity Level}

\newacronym{rcl}{RCL}{ROS Contract Language}

\newacronym{rv}{RV}{Runtime Verification}

\newacronym{rml}{RML}{Runtime Monitoring Language}

\newacronym{sats}{SATS}{Small Aircraft Transportation System}

\newacronym{ico}{ICO}{Interactive Cooperative Objects}

\newacronym{gui}{GUI}{Graphical User Interface}

\newacronym{can}{CAN}{Controller Area Network}

\newacronym{mtbf}{MTBF}{Mean Time Between Failures}

\newacronym{wcet}{WCET}{Worst-Case Execution Time}

\newacronym{caa}{CAA}{Civilian Aviation Authority}

\newacronym{atc}{ATC}{Air Traffic Control}

\newacronym{dlr}{DLR}{German Aerospace Centre}

\glsdisablehyper

\usepackage{tabularx}
\usepackage{textcomp}
\usepackage{wrapfig}
\usepackage{placeins}

\usepackage{zed-csp}

\usepackage[english]{babel}
\usepackage{cleveref}
\crefname{figure}{fig.}{figs.}
\crefname{section}{sect.}{sects.}

\usepackage[disable]{todonotes}

\newcommand{\new}[1]{\textcolor{black}{#1}}

\begin{document}
\maketitle

\begin{abstract}
Software verification is an important tool in establishing the reliability of critical systems.
One potential area of application is in the field of robotics, as robots take on more tasks in both day-to-day areas and highly specialised domains.
Robots are usually given a plan to follow, if there are errors in this plan the robot will not perform reliably.
The capability to check plans for errors in advance could prevent this.
Python is a popular programming language in the robotics domain, through the use of the \gls{ros} and various other libraries.
Python's Turtle package provides a mobile agent, which we formally model here using \gls{csp}.
Our interactive toolchain CSP2Turtle with \gls{csp} model and Python components, enables Turtle plans to be verified in \gls{csp} before being executed in Python.
This means that certain classes of errors can be avoided, and provides a starting point for more detailed verification of Turtle programs and more complex robotic systems.
We illustrate our approach with examples of robot navigation and obstacle avoidance in a 2D grid-world.
\end{abstract}

\glsresetall

\section{Introduction}
\label{sec:intro}

Robotics is a large and complex field, and autonomous robots are often designed to perform essential tasks. 
When it comes to such hybrid safety-critical systems, they demand a high level of confidence in their design and specifications.
Beyond software testing, the highest level of assurance comes from formal verification, and many different approaches have been taken to apply this to robotic systems \cite{luckcuck2019formal}.

Robots are typically given a plan to follow in the form of instructions (``move forward 10m, turn 90\textdegree{}'') or a goal to accomplish (``reach point $A$'').
If this plan asks for something impossible because it does not align with the physical reality of the robot's environment, or if it asks for something logically contradictory, then this plan is incorrect.
This would cause the robot to have incorrect or unpredictable behaviour.
Testing a plan's correctness may be difficult and expensive in both time and money.
Simulations can be time consuming to run, and real world tests may be risky and infeasible.
Checking the plan against the robot's specifications and relevant properties of its environment during the design phase would avoid producing incorrect plans, and avoid the overheads associated with a design loop of producing a plan, running it on the robot, and seeing how the robot performs.

In this paper we present a model of Python's Turtle package in \gls{csp} and show how this can be used to check that a plan for the turtle's behaviour is valid.
We assume the plan is supplied by the user as planning itself encompasses a broad range of activities which are outside the scope of this paper.
We also develop and describe a Python-based tool chain that takes a plan, performs checks on the model, and generates the corresponding Python Turtle code that performs the validated plan.

In the remainder of this section, we provide an overview of both Turtle and \gls{csp}.
Next, \Cref{sec:modelling} describes our approach to modelling and the associated toolchain that was leveraged for verification and code synthesis.
We illustrate our approach via examples and evaluate it in \Cref{sec:example}.
Finally, \Cref{sec:conclude} concludes and identifies future research directions.

\subsection{Turtle}
\label{sec:introTurtle}

Turtle is an in-built Python package\footnote{Turtle package: \url{https://docs.python.org/3/library/turtle.html}} centred around controlling an agent (the ``turtle'') on a 2D plane (the ``canvas'').
It is based on the Logo programming language, which has been used to program physical turtle robots\footnote{Logo Turtle robot: \url{https://web.archive.org/web/20150131192445/https://el.media.mit.edu/logo-foundation/logo/turtle.html}}, and provides a Logo-like set of commands via Python methods.
These produce a graphical display of the turtle following the plan of these commands, tracing a line from its path.
These features make Turtle useful for modelling and testing simple robotics plans and simulating the result.

The turtle has navigation commands for movement (\verb|forward(), backward()|) and changing its direction (\verb|left(), right()|); arguments passed to these functions indicate the distance it should move and the angle it should turn, respectively.
The turtle also has commands controlling a ``pen'' that draws as the turtle moves along its canvas but only if it is in the down position (\verb|penup(), pendown()|).
These six commands are the focus of our modelling approach, described in \Cref{sec:turtle}.

Only a certain amount of the canvas is visible to the user, but the turtle can move and draw outside this.
There are other cosmetic options such as pen colour, pen width, turtle shape, etc.; and additional commands, including leaving a stamp at the turtle's current location, and the ability to start `recording' the turtles movements and draw a polygon along its path when it's finished moving.

\subsection{Communicating Sequential Processes}
\label{sec:introCSP}

\gls{csp} is a formal language for modelling concurrent systems \cite{hoare1978communicating}.
The two fundamental units that \gls{csp} provides are processes and events.
Events are communications over a channel, and a channel's declaration determines if its events take parameters.
Parametrised channels are not used in this model, so none of the events take parameters either.
A specification's behaviour is defined by processes, which themselves are sequences of events.

Events can be combined using various operators, and \gls{csp} provides some in-built processes.
\Cref{tab:cspOperators} summarises the \gls{csp} operators and the two in-built processes ($\Skip, \Stop$) that we use in constructing the model.
The most essential operator to understand is $a \then P$, which defines the process that receives event $a$ then after that acts like the process $P$.
From this it is possible to perform actions such as recursively defining a process.
The other key operators used here are running two process in parallel with no requirements to act in synch on events ($\interleave$), and allowing a process to choose between two or more options as to its next action ($\extchoice$).

The other key concept from \csp employed in this paper is that of a \emph{trace}.
The trace of a process is a sequence of the events that have happened throughout its lifetime.
In our model, these events are designed to be in close correspondence to the functions that would be called by the turtle process in Python, to allow direct translation.

A specification written in \csp can have assertions made about it which are then checked by an appropriate model checker, here the command line tools from the
\gls{fdr} \cite{fdr}, are used.
\gls{fdr} supports \gls{cspm}, which allows for defining parametrised processes in a functional, Haskell-like way.
These assertions are made about the possible traces of the turtle process, in \csp this is known as \emph{trace refinement}.
A process $P$ it said to \emph{trace-refine} another process $Q$ if every (finite) trace of $P$ is also one of $Q$\cite{ucs}.

\begin{table}[t]
\centering
\scalebox{0.9}{\begin{tabularx}{0.9\textwidth}{ l | l | X  }
    \hline
Action			        & Syntax 				& Description \\
\hline \hline
Skip				& $\Skip$ 				& The process that immediately terminates\\
\hline
Stop				& $\Stop$ 				& The process that accepts no events and thus deadlocks\\
\hline
Simple Prefix	                &  $a \then P$		                & Communicate event $a$, then act like process $p$ \\
\hline
External Choice & $P \extchoice Q$		                        & Offer a choice between two processes $P$ and $Q$\\
\hline
Interleaving 	& $P \interleave Q$ 		& Processes $P$ and $Q$ run in parallel with no synchronisation \\
\hline
Hide & $ P \hide A$		                        & The process $P$ runs normally but if any event from set $A$ is performed it is hidden from the trace\\
\hline \hline
\end{tabularx}}
\caption{Summary of \new{the} \gls{csp} operators \new{that are} used in this paper. \label{tab:cspOperators}}
\end{table}

\section{Modelling and Implementation}
\label{sec:modelling}

\subsection{Modelling Approach}
\label{sec:turtle}

We modelled a simplified version of Turtle in \gls{csp}, while still capturing the package's core ideas and functionality.
The turtle is modelled as a \gls{csp} process, and the commands are \gls{csp} events.
The events in our model, and their corresponding turtle commands, are detailed in \Cref{tab:cspEvents}.
The plans we talk about in this paper are comprised of these events.
The first six are abbreviated names of \verb|forward|, \verb|backward|, \verb|penup|, \verb|pendown|, \verb|left|, \verb|right| which were chosen both for brevity and as they are aliases for those methods in Turtle.
We use an additional event $goal$ to mark the location the user has specified in their plan as being the turtle's destination. 

\begin{table}[t]
    \centering
\scalebox{0.9}{\begin{tabularx}{0.95\textwidth}{ l || c | c | c | c | c | c}
        \hline
        CSP Event & $fd$ & $bk$ & $lt$ & $rt$ & $pu$ & $pd$ \\
        \hline
        Turtle command & \texttt{forward()} & \texttt{backward()} & \texttt{left()} & \texttt{right()} & \texttt{penup()} & \texttt{pendown()}
    \end{tabularx}}
    \caption{\new{The} \gls{csp} events \new{that we use in our} model and their corresponding Turtle commands. \label{tab:cspEvents}}
\end{table}

The key simplifications that were made were to restrict the possible directions of the turtle to just the four cardinal directions, and to limit movement to just one unit at a time.
This means the turtle inhabits a grid-world, where its location will always be described by integer co-ordinates.
The decision was also made to bound the size of the world that the turtle inhabits inside of CSP to some user defined constants, and not have it reside in an unbounded world.

These adjustments were made partially for technical reasons to do with our \gls{csp} implementation.
To validate properties of a process, \gls{csp} has to examine all of its states.
This obviously makes infinite state processes impossible to check, which necessitated the bounding of the turtle's grid-world size.
Restrictions on movement and turning were made because \gls{csp} does not natively allow for handling floats, so it would be difficult to place it in a location that was not an integer grid.
Despite these limitations, \gls{csp} still functions as a valuable tool for modelling and examining the properties of the turtle agent, because it allowed the core concepts of movement and pen actions to be formalised.

\subsection{CSP Model Architecture}
\label{sec:arch}

\begin{figure}[t]
    \centering
    \includegraphics[scale=.9]{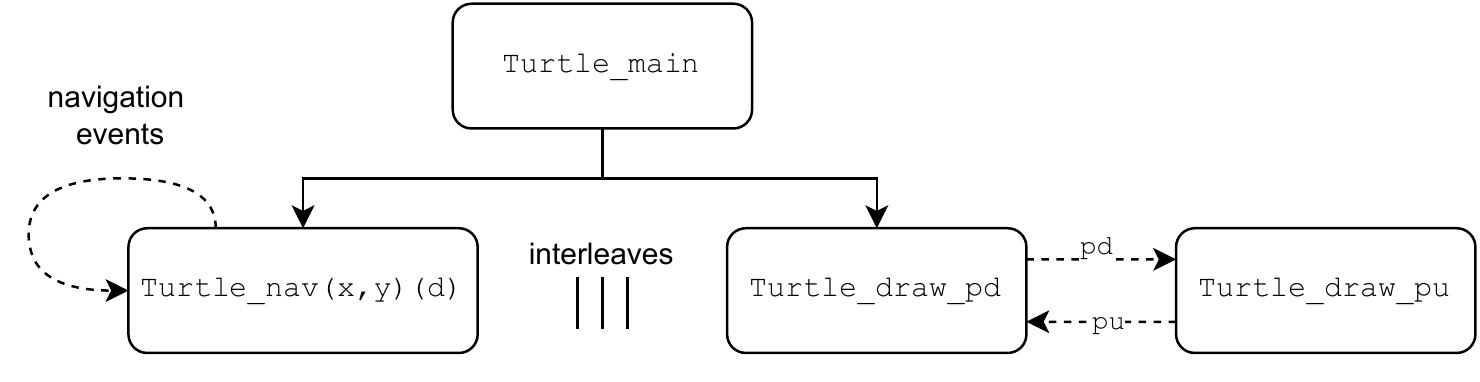}
    \caption{Architecture of the CSP model.
            Each box represents a process. The solid line represents $Turtle\_main$ starting the other processes, which are interleaved $(\interleave)$.
            The dashed arrows ($\dashrightarrow$) indicate events.
            $Turtle\_nav$ responds to the navigation events, which are omitted for brevity.}
    \label{fig:turtlearch}
\end{figure}

\gls{csp} supports specifying complex systems as an interleaving of smaller processes, each specifying one function.
We used this feature in the architecture of our model, as shown in \Cref{fig:turtlearch}.
The \verb|Turtle_main| process is comprised of the \verb|Turtle_nav| and \verb|Turtle_draw_pd| processes running interleaved, it passes the model's initial parameters to each of these more specialised processes.
\verb|Turtle_main| is the process we directly make assertions about, the modular design of composing it out of smaller specialised processes makes it more easily extensible.
The navigation and drawing functionalities of the turtle can be easily separated in this instance as they run independently of each other and do not need to synchronise on any events.
This independence of processes is why the interleave command can be used.
If synchronisation were needed, \csp has a parallel operator that takes two processes and a set of events which they must act in synch on.

\texttt{Turtle\_nav} reacts to the movement and rotation events ($fd$, $bk$, $lt$, and $rt$), and updates the turtle's location ($x, y$) and direction ($d$) that it is currently facing.
To handle these different cases \texttt{Turtle\_nav} uses a different specialised process for each direction.
Our turtle specification contains knowledge of the world's parameters (see \Cref{sec:python}) and \texttt{Turtle\_nav} uses this internal knowledge to avoid moving beyond the boundaries or coming into contact with obstacles.

The \verb|Turtle_draw_pd| process only interacts with the \verb|pu| event, before transitioning to being the opposite process that only waits for a \verb|pd| event.
We assume that the pen always start down, and thus start with the $pd$ process, as this is the default behaviour in Turtle.
This flip-flop system ensures that the trace of this system will always be balanced.
Note that the Turtle package does allow the \texttt{pendown()} method to be called while the pen is already down, but as this has no effect it would always be redundant to have in a plan for a turtle.
This system guarantees such redundancy cannot occur.

\FloatBarrier
\subsection{Python Toolchain}
\label{sec:python}

\begin{figure}[t]
    \centering
    \includegraphics[scale=.8]{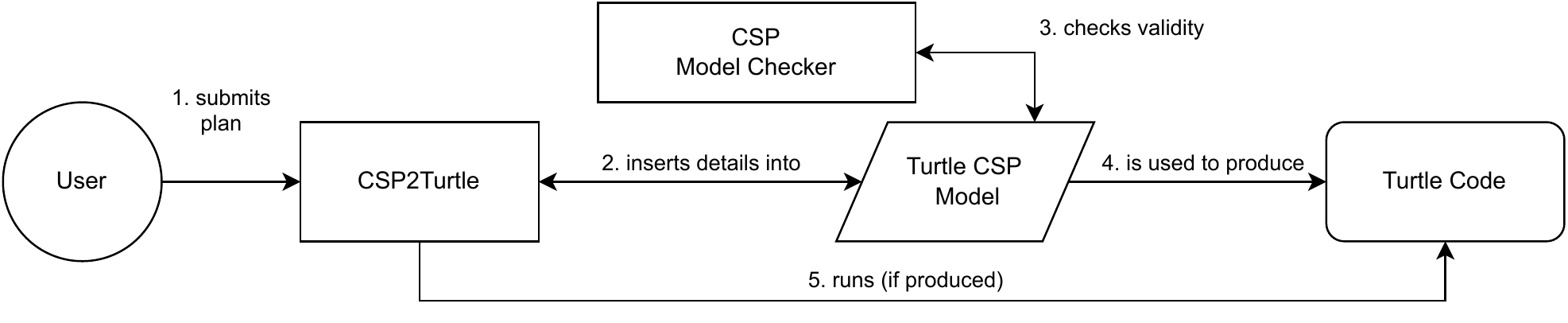}
    \caption{All the stages and components of the toolchain, showing the flow and order of operations, and how the user interacts with it.
            The arrows indicate how components acts on or produce others.
            Circle represents the user, the rectangles are running programs, the parallelogram is the model, and the rounded rectangle is source code produced by the programs.}
    \label{fig:architecture}
\end{figure}

We contribute a prototype toolchain CSP2Turtle to support the use of CSP alongside the Turtle package \footnote{CSP2Turtle code and turtle package model: \url{https://doi.org/10.5281/zenodo.6671802}}.
This Python toolchain is the overall system in which the Turtle CSP model sits and as shown in \Cref{fig:architecture} it is through this toolchain that information is passed in and out.
CSP2Turtle gets world dimensions, a plan, a location in the world to be marked as a goal, and obstacle locations from the user, and synthesises them with the existing \csp model.
It then executes \gls{fdr}'s command line program \verb|refines| to automatically perform two checks on the model, and if a valid plan has been provided it generates and runs Turtle code, giving live graphical output.

The first check is if the plan is valid for the grid-world's size and potential obstacles.
It does this by seeing if the plan corresponds to a possible trace of the system with: 

\centerline{\texttt{assert Turtle_main(0, 0)(E) :[has trace] <plan, as, trace>}.}
\noindent Note that \texttt{Turtle_main} takes its arguments in two pairs of parenthesis.
This allows for it to be partially applied to just the first arguments if necessary, providing programming flexibility.
The second check is if the goal location provided by the user is reachable from the turtle's starting location $(0, 0)$.
The turtle always starts at $(0, 0)$ for consistency, this can be visualised as the bottom left corner of a grid extending up and to the right.
This is a check on the environmental situation rather than the turtle, as it is only concerned with the location of the goal, and potential obstacles between that and the turtle's starting position.
It does not attempt to reach the goal with the trace used in the first assertion.
This check is performed with a trace-refines assertion: 

\centerline{\texttt{assert Turtle_main(0, 0)(E) $\hide$ nav_events [T= goalpoint}.}
\noindent This assertion asks if another process $goalpoint$ trace refines the \texttt{Turtle_main} process.
This goal process has the form \texttt{goalpoint = goal -> STOP} and so only has one possible trace which is a singular goal event.
The $goal$ event can only occur in \texttt{Turtle_main} after it has entered the user defined goal state.
The $\hide nav\_events$ hides the set of navigation events from the trace, meaning the trace will only contain either the goal event or nothing.
This allows for easier checking as we are not concerned with how the turtle reaches the goal, only if it is possible to.
The obstacles that may prevent the turtle from reaching its goal are defined by telling the process to do nothing but $\Skip$ if in the obstacle location.
The results of these checks, pass or fail, are displayed to the user.

If the plan given is a valid trace of the system, a corresponding turtle program will be produced, saved, and ran, with the graphical output being visible to the user.
If the trace is invalid, then no program is produced.
The next section illustrates our approach via example, including code snippets and screenshots from our toolchain.

\section{Example Usage}
\label{sec:example}

\begin{figure}[t]
    \includegraphics[width=0.5\linewidth]{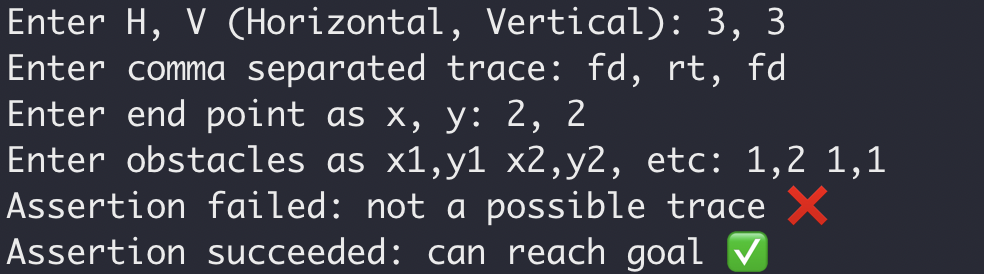}
    \hfill
    \includegraphics[width=0.5\linewidth]{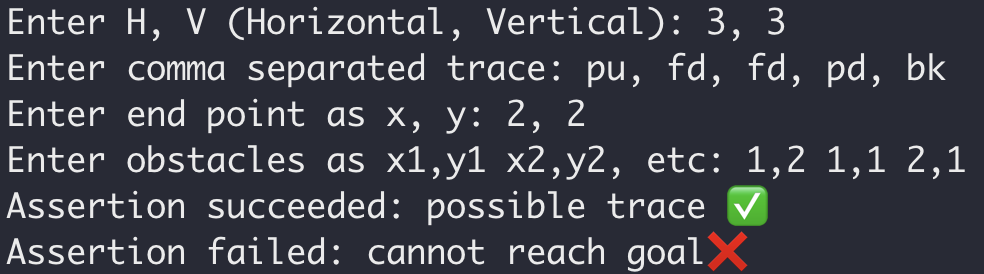}
    \begin{center}
        \includegraphics{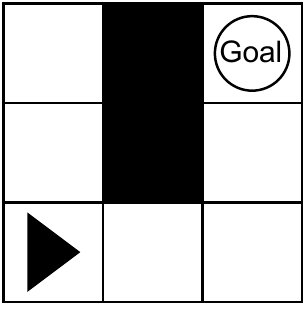}
        \includegraphics{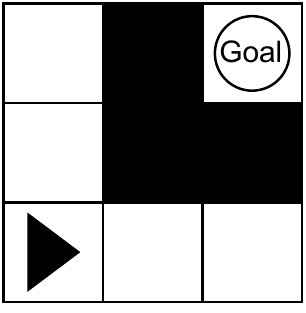}
    \end{center}
    \caption{Two usage examples, with corresponding maps of their worlds beneath them.
    Notice the addition of $(2, 1)$ in the second example blocks the turtle's ($\blacktriangleright$) access to the goal and how the trace in the first example cannot fit in the boundaries.}
    \label{fig:screenshot}
\end{figure}

Our workflow begins with the user initiating the CSP2Turtle prompt.
From here they can input the dimensions of the turtle's world by giving a width and height, and mark certain locations as being ``obstacles'', which are impassable to the turtle.
They can then provide a plan for the turtle in the form of a trace, which \gls{fdr} will check against the parameters of the world to verify if it is a possible plan to execute.
A certain location in the world can be labelled as a goal and \gls{fdr} will deduce if it is reachable from the turtle's starting state.
The starting location is hard-coded to be $(0,0)$ and facing East as this is the default in Turtle, but this could be changed if needed.
The screenshots in \Cref{fig:screenshot} depict examples of CSP2Turtle at runtime and an example map of the turtles' worlds.

The left screenshot shows that the given trace is a not a possible one of that Turtle system.
Turning right then attempting to move forward would move the turtle outside of the confines of its 3x3 world.
Independently of that, the turtle is still able to reach the goal state of (2, 2).
The locations (1, 2) and (1, 1) are blocked as obstacles, but \gls{fdr} finds that it can still move through (2, 1) to reach the goal.

The right screenshot shows a valid trace.
The turtle does not attempt to move beyond the bounds of the world, and performs pen ups and pen downs correctly.
One additional obstacle is added at location (2, 1) and now the turtle can no longer reach the goal from its starting position, \gls{fdr} correctly finding that all entrances are now blocked.
All of this checking takes place before the plan is run, and incorrect plans cannot be run.
This provides a useful `offline' checking mechanism that prevents it being necessary to run a full simulation.

\csp lends itself well to modelling the core Turtle components.
Processes and events map naturally onto the turtle and its methods.
This made the initial modelling decisions easy and provided a good foothold to explore different possibilities for the model.
They were quick to prototype and easy to test and probe in \gls{fdr}.

Adding the Python layer allows for quick and interactive testing of \csp plans.
It also introduces another layer of the system that itself is not verified, which is a weakness when trying to have a highly robust system.
This added layer of complexity enabled tasks that would not be possible in pure \csp, but made it harder to discern if errors were in the Python program or the \csp model.

Other aspects of \csp also influenced design decisions.
The lack of floats meant we decided to not allow arbitrary movement, instead restricting ourselves to the integer grid. 
The way \csp handles variables meant that recording the path of the turtle to incorporate its \texttt{begin\_poly} method would require a much more roundabout approach.
In addition, implementing the plan as a trace is very easy and natural in \csp but other methods would be possible too, such as letting the plan take the form of a process.

\section{Conclusions and Future Work}
\label{sec:conclude}

Robotics can be used to provide safety-critical systems, and such systems require a software verification approach to their design and implementation to ensure that they behave correctly. 
CSP2Turtle enables quick checks on the validity or possibility of certain actions of the robotic agent turtle given certain movement constraints.
\gls{fdr} could be used not just to verify the plans but also to find them.
This work focused only on user submitted plans but extensions like this are certainly an avenue we would like to explore in the future.

Future work includes restructuring the design to capture Turtle's \verb|begin_poly()| and \verb|end_poly()| functions, which facilitate the creation of arbitrary polygons by recording the turtle's movements.
This type of memory-storage driven action is less amenable to implementation in \csp but is still possible.
The scalability of these methods and the core model to larger scale systems could be tested. 

Methods can be developed to alter the topology or geometry of the world.
This could involve changing the topology to objects such a loop or torus, or the geometry to non-rectangular shapes.
An adaptation in a similar vein would be increasing the turtle's range of motion.
Allowing diagonal movement, and a variable move distance would make the model less abstracted
This may require using parametrised channels or allowing for more options but from a fixed range (30\textdegree{}, 45\textdegree{}, etc).
The plans themselves can be made more dynamic too, as mentioned in \Cref{sec:example} a process could be supplied by the user, better representing the types of choices a robot may have in its plans.

The focus of this paper was on a single turtle agent, but the library has the capability for multiple independent agents, which is an area where \csp has been successfully applied before \cite{LuckcuckFormal2021}.
The \csp variant tock CSP could be used to introduce the notion of discrete time into the model, this doesn't correspond to the turtle software but could be useful especially with multiple agents.
Another approach (noted in \Cref{sec:example}) is verifying the Python components themselves.
A way around this would be to use Nagini \cite{chocklernagini2018}, a library for verifying Python code, that can be written inside Python itself instead of having to model it in another external tool.
Nagini could be used to make and prove assertions about the correctness of CSP2Turtle, for instance if it inserts the plan into the model correctly, which would help avoid errors that can occur from adding more programming complexity.

\bibliographystyle{eptcs}
\bibliography{bibliography}

\begin{thebibliography}{1}
\providecommand{\bibitemdeclare}[2]{}
\providecommand{\surnamestart}{}
\providecommand{\surnameend}{}
\providecommand{\urlprefix}{Available at }
\providecommand{\url}[1]{\texttt{#1}}
\providecommand{\href}[2]{\texttt{#2}}
\providecommand{\urlalt}[2]{\href{#1}{#2}}
\providecommand{\doi}[1]{doi:\urlalt{http://dx.doi.org/#1}{#1}}
\providecommand{\eprint}[1]{arXiv:\urlalt{https://arxiv.org/abs/#1}{#1}}
\providecommand{\bibinfo}[2]{#2}

\bibitemdeclare{incollection}{chocklernagini2018}
\bibitem{chocklernagini2018}
\bibinfo{author}{Marco \surnamestart Eilers\surnameend} \&
  \bibinfo{author}{Peter \surnamestart Müller\surnameend}
  (\bibinfo{year}{2018}): \emph{\bibinfo{title}{Nagini: {A} {Static} {Verifier}
  for {Python}}}.
\newblock In \bibinfo{editor}{Hana \surnamestart Chockler\surnameend} \&
  \bibinfo{editor}{Georg \surnamestart Weissenbacher\surnameend}, editors: {\sl
  \bibinfo{booktitle}{Computer {Aided} {Verification}}},
  \bibinfo{volume}{10981}, \bibinfo{publisher}{Springer International
  Publishing}, \bibinfo{address}{Cham}, pp. \bibinfo{pages}{596--603},
  \doi{10.1007/978-3-319-96145-3_33}.
\newblock
  \urlprefix\url{http://link.springer.com/10.1007/978-3-319-96145-3_33}.

\bibitemdeclare{inproceedings}{fdr}
\bibitem{fdr}
\bibinfo{author}{Thomas \surnamestart Gibson-Robinson\surnameend},
  \bibinfo{author}{Philip \surnamestart Armstrong\surnameend},
  \bibinfo{author}{Alexandre \surnamestart Boulgakov\surnameend} \&
  \bibinfo{author}{Andrew \surnamestart Roscoe\surnameend}
  (\bibinfo{year}{2014}): \emph{\bibinfo{title}{{FDR3 -- A Modern Model Checker
  for CSP}}}.
\newblock In: {\sl \bibinfo{booktitle}{Tools and Algorithms for the
  Construction and Analysis of Systems}}, {\sl \bibinfo{series}{LNCS}}
  \bibinfo{volume}{8413}, \bibinfo{publisher}{Springer}, pp.
  \bibinfo{pages}{187--201}, \doi{10.1007/978-3-642-54862-8\_13}.

\bibitemdeclare{article}{hoare1978communicating}
\bibitem{hoare1978communicating}
\bibinfo{author}{Charles Antony~Richard \surnamestart Hoare\surnameend}
  (\bibinfo{year}{1978}): \emph{\bibinfo{title}{Communicating sequential
  processes}}.
\newblock {\sl \bibinfo{journal}{Communications of the ACM}}
  \bibinfo{volume}{21}(\bibinfo{number}{8}), pp. \bibinfo{pages}{666--677},
  \doi{10.1145/359576.359585}.

\bibitemdeclare{inproceedings}{LuckcuckFormal2021}
\bibitem{LuckcuckFormal2021}
\bibinfo{author}{Matt \surnamestart Luckcuck\surnameend} \&
  \bibinfo{author}{Rafael~C. \surnamestart Cardoso\surnameend}
  (\bibinfo{year}{2022}): \emph{\bibinfo{title}{Formal Verification of a Map
  Merging Protocol in the Multi-agent Programming Contest}}.
\newblock In \bibinfo{editor}{Natasha \surnamestart Alechina\surnameend},
  \bibinfo{editor}{Matteo \surnamestart Baldoni\surnameend} \&
  \bibinfo{editor}{Brian \surnamestart Logan\surnameend}, editors: {\sl
  \bibinfo{booktitle}{Engineering Multi-Agent Systems}},
  \bibinfo{publisher}{Springer}, \bibinfo{address}{Cham}, pp.
  \bibinfo{pages}{198--217}, \doi{10.1007/978-3-030-97457-2_12}.
\newblock \urlprefix\url{https://arxiv.org/abs/2106.04512}.

\bibitemdeclare{article}{luckcuck2019formal}
\bibitem{luckcuck2019formal}
\bibinfo{author}{Matt \surnamestart Luckcuck\surnameend},
  \bibinfo{author}{Marie \surnamestart Farrell\surnameend},
  \bibinfo{author}{Louise~A \surnamestart Dennis\surnameend},
  \bibinfo{author}{Clare \surnamestart Dixon\surnameend} \&
  \bibinfo{author}{Michael \surnamestart Fisher\surnameend}
  (\bibinfo{year}{2019}): \emph{\bibinfo{title}{Formal specification and
  verification of autonomous robotic systems: A survey}}.
\newblock {\sl \bibinfo{journal}{ACM Computing Surveys (CSUR)}}
  \bibinfo{volume}{52}(\bibinfo{number}{5}), pp. \bibinfo{pages}{1--41},
  \doi{10.1145/3342355}.
\newblock \urlprefix\url{https://dl.acm.org/doi/10.1145/3342355}.

\bibitemdeclare{book}{ucs}
\bibitem{ucs}
\bibinfo{author}{A.W. \surnamestart Roscoe\surnameend}:
  \emph{\bibinfo{title}{Understanding Concurrent Systems}}.
\newblock \bibinfo{series}{Texts in Computer Science},
  \bibinfo{publisher}{Springer London}, \doi{10.1007/978-1-84882-258-0}.
\newblock \urlprefix\url{http://link.springer.com/10.1007/978-1-84882-258-0}.

\end{thebibliography}
\end{document}